\documentclass[11pt]{article}

\newcommand{\ours}{$\mathsf{GLAINTEL}$}

\usepackage[final]{acl}

\usepackage{times}
\usepackage{latexsym}

\usepackage[T1]{fontenc}

\usepackage[utf8]{inputenc}

\usepackage{microtype}

\usepackage{inconsolata}

\usepackage{graphicx}

\title{Grounded Language Agent for Product Search via Intelligent Web Interactions}

\author{Moghis Fereidouni, Adib Mosharrof, A.B. Siddique \\
  University of Kentucky, Lexington, KY, USA\\
  \texttt{moghis.fereidouni@uky.edu}, \texttt{amo304@g.uky.edu}, \texttt{siddique@cs.uky.edu}  \\}

\usepackage{hyperref}
\usepackage{enumitem}
\usepackage{graphicx}
\usepackage{subcaption}
\usepackage{multirow}
\usepackage{booktabs} 
\usepackage{balance}
\usepackage{threeparttable}
\usepackage{amssymb}
\usepackage{amsmath}
\usepackage{verbatim}
\usepackage{graphicx}

\begin{document}

\newcommand{\cmt}[1]{{\footnotesize\textcolor{red}{#1}}}
\newcommand{\cmto}[1]{{\footnotesize\textcolor{orange}{#1}}}
\newcommand{\note}[1]{\cmt{Note: #1}}
\newcommand{\TODO}[1]{\cmt{TO-DO: #1}}
\newcommand{\question}[1]{\cmto{Question: #1}}
\newcommand{\abu}[1]{{\footnotesize\textcolor{blue}{Abu: #1}}}
\newcommand{\edits}[1]{\textcolor{red}{#1}}
\newcommand{\myNum}[1]{(\emph{#1})}
\newcommand{\myAbc}[1] {($\mathtt{#1}$)}

\newcommand{\x}{\mathbf{x}}
\newcommand{\z}{\mathbf{z}}
\newcommand{\y}{\mathbf{y}}
\newcommand{\w}{\mathbf{w}}
\newcommand{\data}{\mathcal{D}}

\newcommand{\etal}{{et~al.}\ }
\newcommand{\eg}{e.g.,\ }
\newcommand{\ie}{i.e.,\ }
\newcommand{\nth}{\text{th}}
\newcommand{\pr}{^\prime}
\newcommand{\tr}{^\mathrm{T}}
\newcommand{\inv}{^{-1}}
\newcommand{\pinv}{^{\dagger}}
\newcommand{\real}{\mathbb{R}}
\newcommand{\gauss}{\mathcal{N}}
\newcommand{\norm}[1]{\left|#1\right|}
\newcommand{\trace}{\text{tr}}

\newcommand{\reward}{r}
\newcommand{\policy}{\pi}
\newcommand{\mdp}{\mathcal{M}}
\newcommand{\states}{\mathcal{S}}
\newcommand{\actions}{\mathcal{A}}
\newcommand{\observations}{\mathcal{O}}
\newcommand{\transitions}{\mathcal{T}}
\newcommand{\initstate}{d_0}
\newcommand{\freq}{d}
\newcommand{\obsfunc}{E}
\newcommand{\initial}{\mathcal{I}}
\newcommand{\horizon}{H}
\newcommand{\goal}{\mathcal{G}}
\newcommand{\rewardevent}{\mathcal{R}}
\newcommand{\probr}{p_\rewardevent}
\newcommand{\metareward}{\bar{\reward}}
\newcommand{\discount}{\gamma}
\newcommand{\behavior}{{\pi_\beta}}
\newcommand{\bellman}{\mathcal{B}}
\newcommand{\qparams}{\phi}
\newcommand{\qparamset}{\Phi}
\newcommand{\qset}{\mathcal{Q}}
\newcommand{\batch}{B}
\newcommand{\qfeat}{\mathbf{f}}
\newcommand{\Qfeat}{\mathbf{F}}
\newcommand{\vocabulary}{\mathcal{V}}

\newcommand{\traj}{\tau}

\newcommand{\pihi}{\pi^{\text{hi}}}
\newcommand{\pilo}{\pi^{\text{lo}}}
\newcommand{\ah}{\mathbf{w}}

\newcommand{\proj}{\Pi}

\newcommand{\loss}{\mathcal{L}}
\newcommand{\eye}{\mathbf{I}}

\newcommand{\model}{\hat{p}}

\newcommand{\pimix}{\pi_{\text{mix}}}

\newcommand{\pib}{\bar{\pi}}
\newcommand{\epspi}{\epsilon_{\pi}}
\newcommand{\epsmodel}{\epsilon_{m}}

\newcommand{\return}{\mathcal{R}}

\newcommand{\conpen}{\mathcal{C}}

\newcommand{\cY}{\mathcal{Y}}
\newcommand{\cX}{\mathcal{X}}
\newcommand{\en}{\mathcal{E}}
\newcommand{\be}{\mathbf{e}}
\newcommand{\by}{\mathbf{y}}
\newcommand{\bx}{\mathbf{x}}
\newcommand{\bz}{\mathbf{z}}
\newcommand{\bo}{\mathbf{o}}
\newcommand{\bs}{\mathbf{s}}
\newcommand{\ba}{\mathbf{a}}
\newcommand{\ot}{\bo_t}
\newcommand{\st}{\bs_t}
\newcommand{\at}{\ba_t}
\newcommand{\op}{\mathcal{O}}
\newcommand{\opt}{\op_t}
\newcommand{\kl}{D_\text{KL}}
\newcommand{\tv}{D_\text{TV}}
\newcommand{\ent}{\mathcal{H}}

\newcommand{\bzhi}{\bz^\text{hi}}

\newcommand{\E}{\mathbb{E}}
\newcommand{\dataset}{\mathcal{D}}

\newcommand{\myvalue}[1] {$\mathtt{#1}$}
\newcommand{\myspecial}[1] {\texttt{#1}}

\newcommand{\xx}{\mathbf{x}}
\newcommand{\kk}{\mathbf{k}}

\definecolor{cmarkcolor}{RGB}{21, 164, 64} 
\definecolor{xmarkcolor}{RGB}{177, 0, 4} 

\newcommand{\cmark}{\color{cmarkcolor}\ding{51}}%
\newcommand{\xmark}{\color{xmarkcolor}\ding{55}}%
\newcommand{\mobilerecdataset}{$\mathsf{MobileRec}$}

\newcommand{\stitle}[1]{\noindent\textup{\textbf{#1}}}

\newcommand{\bigX} {\mathcal{X} }
\newcommand{\bigY} {\mathcal{Y} }

\maketitle

\begin{abstract}

The development of agents powered by large language models (LLMs) to accomplish complex high-level user intents, has attracted significant attention recently.
However, employing LLMs with billions of
parameters (e.g., GPT-4) may incur substantial costs
on top of handcrafting extensive prompts.
To address this, we introduce a Grounded Language Agent for Intelligent Web Interactions, named {\ours}. 
{\ours} employs Flan-T5 as its backbone and is flexible in training in various settings: unsupervised learning, supervised learning, and unsupervised domain adaptation.
Specifically, we tackle both the challenge of learning without human demonstrations and the opportunity to leverage human demonstrations effectively when those are available.
Additionally, we explore unsupervised domain adaptation for cases where demonstrations are limited to a specific domain.
Experimental evaluations across diverse setups demonstrate the effectiveness of {\ours} in unsupervised settings, outperforming in-context learning-based approaches that employ larger models with up to 540 billion parameters.
Surprisingly, behavioral cloning-based methods that straightforwardly use human demonstrations do not outperform unsupervised variants of {\ours}. Additionally, we show that combining human demonstrations with reinforcement learning-based training yields results comparable to methods utilizing GPT-4.
The code is available at: \href{https://github.com/MultifacetedNLP/Web-Agents-Unsupervised}{https://github.com/MultifacetedNLP/Web-Agents-Unsupervised}.
\end{abstract}

\section{Introduction}

\begin{figure*}[t!]
  \centering
  \includegraphics[width=0.969\linewidth]{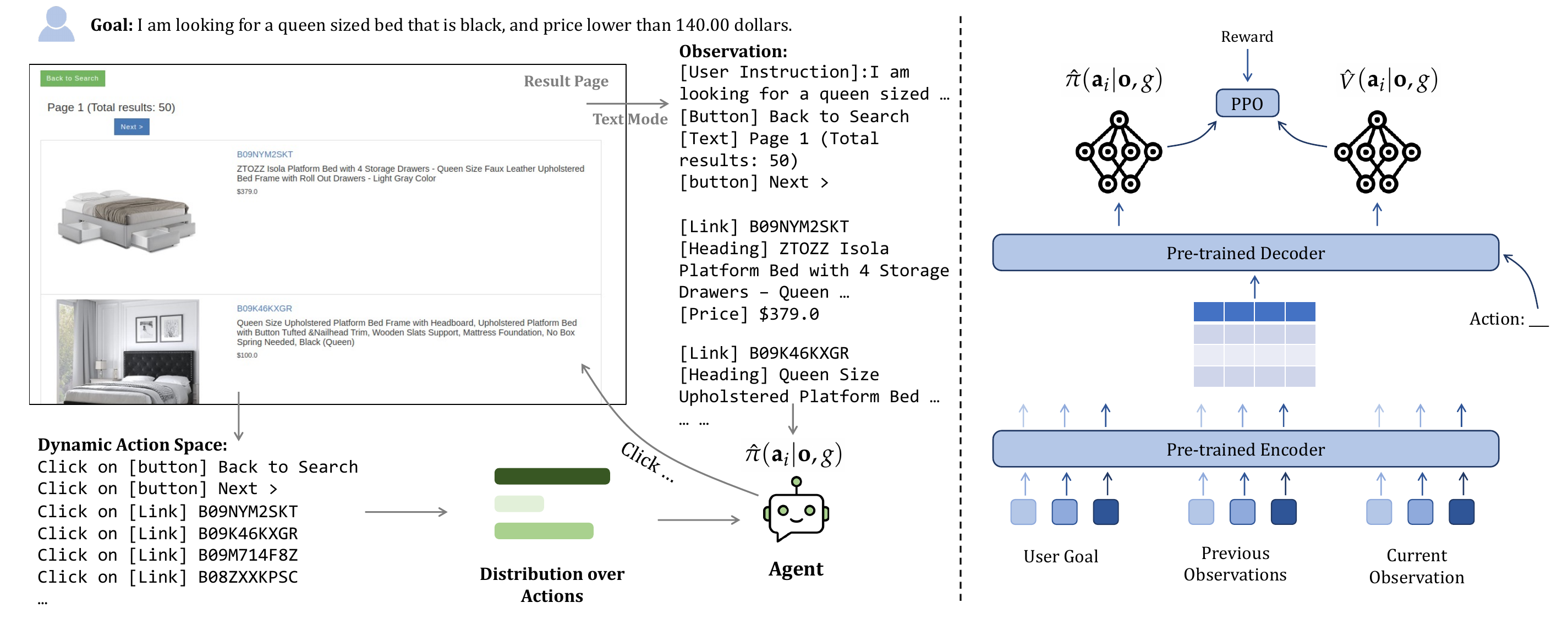}
  \vspace{-6pt}
  \caption{Overview of {\ours}: 
  Our agent employs the Flan-T5 architecture and incorporates a language modeling head to adapt to dynamic action space, while the value head enables precise value estimation.
  }
  \vspace{-6pt}
  \label{fig:intro}
\end{figure*}

Large Language Models~(LLMs) have demonstrated their proficiency in diverse tasks such as text classification, information extraction, and question answering~\cite{bommasani2021opportunities,brown2020language,vaswani2017attention,raffel2019exploring,radford2019language}.
Similarly, reinforcement learning~(RL) has evolved as a powerful paradigm for training intelligent agents to navigate complex environments~\cite{huang2022inner,ahn2022can,liang2023code}.
Moreover, recent research highlights the capabilities of agents powered by LLMs.
For example, agents utilizing GPT-4 can explore the virtual world in Minecraft, acquire a diverse set of composable skills, and exhibit exceptional proficiency in playing the game~\cite{wang2023voyager}.
The exceptional amount of world knowledge, often derived from vast text datasets, opens up possibilities for developing LLM-assisted intelligent web navigation agents capable of navigating and interacting with web pages akin to humans.

Despite their remarkable capabilities, off-the-shelf pre-trained LLMs face challenges in grounding and aligning themselves in interactive web environments~\cite{mahowald2023dissociating}.
This limitation hampers their functional competence without additional customization.
Additionally, employing LLMs with billion-scale parameters, such as GPT-4, may incur substantial costs on top of handcrafting extensive prompts.
On the other hand, training smaller LLMs (e.g., Flan-T5) as agents can be challenging.
For instance, consider a real-world product search scenario, where effective query formulation requires the agent to operate over a huge action space (\ie language vocabulary), and navigating through diverse web pages poses additional challenges that need strategic exploration due to the presence of different actions on each page (\ie dynamic action space).
This complexity prevents the straightforward utilization of an action head on top of LLM.
Moreover, the challenge extends to preserving long-term memory capabilities, which are crucial for comparing items or backtracking during the search process.

In this work, we introduce {\ours}, a \textbf{G}rounded \textbf{L}anguage \textbf{A}gent designed for \textbf{Intel}ligent Web Interactions. 
Given a user's intent specifying a product requirement, {\ours} formulates queries, navigates diverse web pages, and executes various actions to identify, customize, and purchase the desired product.
{\ours} uses the open-source Flan-T5 language model (\ie 780 million parameters) as its backbone and can be flexibly trained in various scenarios: unsupervised, supervised, and unsupervised domain adaptation settings.
Specifically, we address the following research questions.

\begin{itemize}[leftmargin=1.2\parindent,labelindent=-1pt, itemsep=-1pt]

 \item \emph{RQ1: Effectiveness of Unsupervised Learning:} Can LLM-based agents learn to address effective query generation and exploration of complex web pages with no human demonstrations? 

\item \emph{RQ2: Impact of Human Demonstrations:} Can incorporating human demonstrations facilitate LLM-based agents to improve their overall performance? How to effectively leverage human demonstrations for training robust agents?

\item \emph{RQ3: Unsupervised Domain Adaptation:} Can LLM-based agents generalize to new, unseen product categories where no human demonstrations are available? 
\end{itemize}


%

We employ a language modeling head to accommodate a \emph{dynamic action space} and introduce an additional value head for precise value estimates.
Figure~\ref{fig:intro} provides an overview of {\ours}.
The user's goal and observation are sequentially passed to the model at each step.
First, we obtain the input representation for every potential action token and compute the normalized joint probability for each action conditioned on the user goal and observation.
Following the estimation of each action's probability, we apply a softmax function over these probabilities and sample an action according to this distribution.
We fine-tune the agent using the Proximal Policy Optimization~(PPO) algorithm~\cite{baselines}.

We conduct extensive experimental evaluations across diverse setups using the WebShop environment~\cite{NEURIPS2022_82ad13ec}.
WebShop is a simulated yet realistic e-commerce web platform featuring 1.18 million real-world products and 12,087 crowd-sourced natural language intents.
Based on our empirical study, we demonstrate that training Flan-T5 (\eg 780 million parameters) in the unsupervised setting (\ie no human demonstrations) can outperform in-context learning methods~\cite{sridhar2023hierarchical} that rely on models with up to 540 billion parameters.
To quantify the impact of human supervision, we utilized 1010 human demonstrations for training supervised learning models using behavior cloning~(BC)~\cite{pomerleau1989alvinn}.

Our findings indicate that incorporating human demonstrations through \emph{straightforward BC does not produce superior results} when compared to the unsupervised RL-based PPO algorithm.
Furthermore, our investigations reveal that leveraging human demonstrations through BC and then further training the agent with PPO in the unsupervised setting leads to the best results.
Remarkably, this approach achieves results comparable to the method~\cite{ma2023laser} that utilizes GPT-4.
In the unsupervised domain adaptation~(UDA) experiment, we observe that incorporating human demonstrations from a single category enables the agent to generalize to new product categories where no human demonstrations are available.
Additionally, we \emph{evaluate our trained model on a real website eBay} without any additional fine-tuning, which shows comparable results to methods that use the state-of-the-art GPT-4 model.

\section{Proposed Agent: {\ours}}

\subsection{Problem Formulation}
Given a user intent in natural language, the agent's goal is to buy the most appropriate product that fulfills the user's intent.
We formulate the task as a goal-augmented Partially Observable Markov Decision Process $\mdp = (\states,\actions,\transitions,\rewardevent,\goal,\observations,\gamma)$, where $\states$ is a set of states $\bs \in \states$;
$\actions \subset \vocabulary^N$ represents action space sampled from LLM's vocabulary $\vocabulary$ of size $N$;
$\goal \subset \vocabulary^N$ denotes the goal space;
$\transitions: \states \times \actions \mapsto \states$ is the transition function;
$\rewardevent: \states \times \actions \times \goal \mapsto \real$ characterizes the goal-conditioned reward function;
$\observations$ is a set of observations $\bo \in \observations$ (\ie~web page visible to agent);
$\gamma$ is the discount factor.
We employ the language modeling head (\ie distribution over the vocabulary) to accommodate the dynamic action space, which also facilitates directly computing the log probabilities of each action $\ba_i = ( w_0, \cdots, w_{|\ba_i|} )$ sampled from a dynamic action space given the agent's goal $g \in \goal$ and observation $\bo$.

It is important to note that each observation (\ie~web page) presents a dynamic set of actions to the agent, which prevents us from learning a probability distribution over the action space as in classification tasks.
For instance, a search page allows actions such as typing an open-ended textual query or pressing the `Search' button.
Conversely, a product detail page offers actions such as `Back to Search', `< Prev', `Description', `Features', `Reviews', `Buy Now', and the product-specific variable number of options.
Figure~\ref{fig:intro} shows the observation and action space for the `search result page'.

\subsection{Overview of {\ours}}

We employ Flan-T5~\footnote{Checkpoints: \href{https://github.com/google-research/t5x/blob/main/docs/models.md\#flan-t5-checkpoints}{https://github.com/google-research/t5x/ blob/main/docs/models.md\#flan-t5-checkpoints}} as the core architecture, with the integration of the language modeling head and value head on top of the model.
Our proposed agent, {\ours}, is adaptable to training across various setups: 
\myNum{i}~unsupervised learning: no human demonstrations are available;
\myNum{ii}~unsupervised domain adaptation: limited human demonstrations in a single domain are available; and 
\myNum{iii}~supervised learning: human demonstrations are accessible.
In the following, we detail the specifics of the training and inference phases.
The inclusion or exclusion of these phases is contingent upon the availability of the human demonstration data.

\subsection{Optional Phase One: Supervised Training}

The human demonstrations can serve as mappings from states to actions. 
Techniques such as imitation learning or behavioral cloning (BC)~\cite{pomerleau1989alvinn} can be employed to fine-tune the policy $\pi$ by minimizing the following loss over a dataset $\data$ comprising human demonstrations:
\vspace{-6pt}
\begin{equation*}
\mathcal{L}(\pi) = \E_{(s,a)\sim \data}[-\log\pi(a|s)].
\end{equation*}
\vspace{-16pt}

The above formulation can be adapted to incorporate the interaction history with web pages $\pi(\ba_t|\bs_t,\tau_{<t})$, where $\tau_{<t}$ refers to the interaction trajectory leading up to time $t$.
Subsequently, this formulation readily extends to utilize LLMs to learn an optimal policy where the encoder encodes the history of observations $(\bs_t, \tau_{<t})$ and the decoder generates the next action $\ba_t$ as:
\vspace{-6pt}
\begin{equation*}
\mathcal{L}_\text{LLM}(\pi) = \E_{\tau\sim\data}[\sum_{t=0}^L-\log\pi(\ba_t|\tau_{<t},\bs_t)].
\end{equation*}
\vspace{-12pt}

Building upon the recent works in return-conditioned supervised learning~\cite{brandfonbrener2022does,paster2022you,yang2022dichotomy}, we introduce an additional conditioning variable $g \in \goal$ (\ie user goal).
This variable captures overall trajectory-level information, to steer the model toward the goal.
Moreover, in implementation, we use observations $\bo$ (\ie visible web page) instead of the actual state $\bs$. 
Our final formulation is expressed as:
\vspace{-6pt}
\begin{equation*}
\mathcal{L}_\mathrm{LLM}(\pi) = \E_{\tau\sim\data}[\sum_{t=0}^L -\log \pi (\ba_t|\tau_{<t}, \bo_t, g) ].
\end{equation*}
\vspace{-12pt}

The training of this phase can be skipped or chosen based on the availability and feasibility of acquiring human demonstrations.
In our approach to address RQ1~(Effectiveness of Unsupervised Learning), we skip this phase.
We limit the human demonstration data to a single category for RQ3~(Unsupervised Domain Adaptation).
To investigate RQ2~(Impact of Human Demonstrations), we utilize all the available training data for the supervised training phase.

\subsection{Phase Two: Unsupervised Training}

The unsupervised learning phase, which forms the core of the proposed agent {\ours}, operates without any human demonstrations.
This phase is designed to autonomously learn and adapt without relying on expert-guided examples.
The objective of the agent is to learn a policy $\pi: \observations \times \goal \mapsto \mathbb{P}(\mathcal{A})$ that optimizes the expected discounted cumulative rewards for a given goal $g$.
In this work, we leverage PPO algorithm for training, which simultaneously learns a policy $\hat{\pi}$ and a value function $\hat{V}: \observations \times \goal \mapsto \real$ approximating to the true value $V(\bs, g) = \E_{\ba \sim \hat{\pi}(\observations(\bs),g)}\bigl[\return(\bs,\ba,g) + \gamma V(\transitions(\bs,\ba),g)\bigr]$.
We can calculate the probability of each action $\ba_i \in \actions$ using the likelihood computed by the model, expressed as:
$\hat{\pi}(\ba_i|\bo,g) = P(\ba_i|g)$.
That is, the likelihood of choosing each action is calculated based on the probability distributions associated with the tokens that make up the action.
This approach ties the action probabilities directly to the distributions of the individual tokens involved in constructing the action.
Following~\cite{carta2023grounding}, we incorporate a multilayer perception~(MLP) with a single output on top of the last layer of the model to approximate the value $V$. 
Specifically, we employ the language modeling head to directly compute the log probabilities of each action $\ba_i = \{w_0, \cdots, w_{|\ba_i|}\}$ from the dynamic action space given the agent's goal $g \in \goal$ and observation $\bo_t$ at time $t$ as follows:
\vspace{-8pt}
\begin{equation*}
P(\ba_i) = \frac{1}{|a_i|} \sum^{|a_i|}_{k=0} \log P_{\texttt{LM-head}}(w_k | g, \bo_t, w_{<k}).
\end{equation*}
\vspace{-8pt}


Subsequently, employing the softmax operation, we calculate a probability distribution over the action space $\actions$ as follows:
\vspace{-8pt}
\begin{equation*}
P(\ba_i|g) = \frac{e^{P(\ba_i)}}{\sum_{\ba_k \in \actions} e^{P(\ba_k)}}.
\end{equation*}
\vspace{-8pt}

While the actions comprise multiple tokens, the number of possible actions can vary substantially depending on the current observation (\ie web page), which introduces additional complexity.
This phase is mandatory regardless of whether training is conducted in the optional first phase.

\subsection{Phase Three: Inference}

In the inference phase, various decoding techniques for action selection can be employed, such as greedy decoding and top-p.
Given the well-established nature of these techniques, we omit details and provide key insights only.
Greedy decoding, chosen for action selection, has a drawback as it tends to trap the agent in loops, ultimately resulting in suboptimal overall performance.
Conversely, opting for top-p sampling can yield a higher success rate, as it provides a theoretical tradeoff between sampling and greedy decoding. 
However, the process of determining the optimal values for p can be time-intensive.
To address these issues, we turn to the Epsilon-Greedy algorithm for action selection during inference.
In particular, at a step $t$, the greedy will choose the action with the highest probability, while the epsilon will sample based on the probability distribution across the action space. 
This method achieves a higher success rate and an enhanced overall performance, all while avoiding the issue of getting stuck in loops.
It is worth noting that a judiciously chosen, small value for epsilon has been employed in our work, eliminating the need for an exhaustive search.

\section{Experimental Setup}

\subsection{WebShop Environment}

Webshop~\cite{NEURIPS2022_82ad13ec} is a simulated web-based interactive environment with 1.18 million real-world products and 12,087 crowd-sourced text instructions. 
The goal of the agent is to buy a product with specific attributes and options given natural language instruction.
The environment contains 5 different categories, which exhibit significant dissimilarities, particularly in terms of possessing nearly exclusive attributes.
For instance, as illustrated in Table~\ref{webshop_statistics}, a substantial 95.9\% of Fashion's attributes are unique to its category.

\stitle{Human Demonstrations.}
The Webshop also contains a human demonstration dataset.
The human demonstration dataset encompasses a total of 1010 distinct trajectories, distributed across categories.
This dataset is created by asking humans to demonstrate how they would query a product and then take different steps in the Webshop environment to buy a product with desired options and attributes.

{\ours} has the flexibility to incorporate human demonstrations through optional phase one training. 
We utilize human demonstration data to quantify the impact of human demonstrations (RQ2) and explore UDA (RQ3). 
Additionally, {\ours} can be trained without any human demonstrations (RQ1).


\subsection{Evaluation Methodology}

\stitle{Reward.}
We assign a reward $r \in [0,1]$ to the agent after it completes a purchase at the concluding step of an episode.
Specifically, the reward is determined by how closely the purchased product matches the specific attributes and options mentioned in the user instructions as follows:
\vspace{0.3cm}

\noindent\vspace{0.5cm}\resizebox{\columnwidth}{!}{
$
r = r_{\text{type}} \cdot \frac{ |U_{\text{att}} \cap Y_{\text{att}}| + |U_{\text{opt}} \cap Y_{\text{opt}}| +  1[y_{\text{price}} \leq u_{\text{price}}]}{|U_{\text{att}}| + |U_{\text{opt}}| + 1}
$
}
The reward incorporates three main components: $U_{\text{att}}$, $U_{\text{opt}}$, and $u_{\text{price}}$, representing a set of attributes, a set of options, and the price set down in the user's instruction, respectively. 
Correspondingly, $Y_{\text{att}}$, $Y_{\text{opt}}$, and $y_{\text{price}}$ denote the set of attributes, the set of options, and the actual price of the purchased product by the agent. 
Additionally, $r_{\text{type}}$ functions as a text-matching heuristic, assigning a lower reward when the purchased product and the targeted product in the user instruction have similar attributes and options while being different types of products.
Interested readers are referred to WebShop~\cite{NEURIPS2022_82ad13ec} for details.

\stitle{Evaluation Metrics.} Two evaluation metrics are computed using the rewards obtained from the episodes: \myNum{i}~the Score and \myNum{ii}~the Success Rate.
The Score metric represents the average reward across all test episodes multiplied by 100, while the Success rate metric measures the percentage of test episodes in which the full reward (1 out of 1) was attained.
Given that our inference step incorporates sampling, the reported Score and Success Rate metrics are averaged by running the model four times.
\emph{We provide additional implementation details in  Appendix~\ref{sec:implementation_details}.}

\begin{table}[t!]
\centering
\small
\resizebox{\columnwidth}{!}{%
\begin{tabular}{l|c|c|c}
\toprule
Category & \# Attributes & \begin{tabular}[c]{@{}c@{}}\% Unique \\Attributes  \end{tabular}& \begin{tabular}[c]{@{}c@{}}\# Human\\ Demonstrations \end{tabular}  \\ \hline
Beauty            & 143                    & 85.3\% & 224 \\
Garden            & 133                    & 87.2\% & 211 \\
Grocery           & 117                    & 92.3\% & 189 \\
Electronics       & 141                    & 91.4\% & 169 \\
Fashion           & 173                    & 95.9\% & 217 \\
\bottomrule
\end{tabular}
}
\vspace{-8pt}
\caption{Detail about Webshop Environment.}
\label{webshop_statistics}
\vspace{-16pt}
\end{table}

\subsection{Competing Methods}\label{sec:competingMethods}

\stitle{WebShop Baselines}~\cite{NEURIPS2022_82ad13ec}: 
We consider the following baselines from the WebShop paper: 
\myNum{i}~rule-based (Rule$_{ws}$), 
\myNum{ii}~behavioral cloning-based supervised learning (BC$_{ws}$), \myNum{iii}~two reinforcement learning models—one with a transformer text encoder (PG$_{ws}$) and another with an RNN (RNN$_{ws}$), and
\myNum{iv}~a hybrid method (BC + PG). Human experts (Human) also set a benchmark for human-level performance.




\stitle{DRRN}~\cite{he-etal-2016-deep}: 
DRRN is a classic RL baseline that uses separate neural networks to embed states and actions into embedding vectors. An interaction function (\eg inner product) then computes the Q-function value for the state-action pair.

\stitle{Act and ReAct}~\cite{yao2023react}: 
The ReAct method is an in-context learning approach using LLMs that combines reasoning and action execution to tackle diverse tasks. In the WebShop environment, ReAct adds reasoning at each step to guide the agent's decisions on exploration, purchasing, and option selection.


\stitle{WebGUM}~\cite{furuta2024multimodal}:
WebGUM is an instruction-finetuned model, that is further trained on human demonstrations for web navigation.

\stitle{ASH Prompting}~\cite{sridhar2023hierarchical}: 
ASH consists of two main components: \myNum{i}~ Summarizer condenses observations by retaining only relevant information, and \myNum{ii}~Actor uses this condensed observation to generate the next action.


\stitle{PIX2ACT}~\cite{shaw2023pixels}: PIX2ACT builds upon the Pix2Struct model~\cite{Lindenberger_2021_ICCV}, utilizing an image transformer encoder along with a text transformer decoder.

\stitle{LASER}~\cite{ma2023laser}:
LASER is a GPT-4-based method that converts an interactive decision-making task into state space exploration by mapping all possible observations to a finite set of states, with the agent navigating these states through predefined actions specific to each state

\stitle{Prospector}~\cite{kim2023prospector}:
The Prospector uses two approaches: the AskAct method, which incorporates self-asking steps in few-shot demonstrations to extract actions from LLMs, and the Trajectory Ranking (TR) method, where LLMs generate diverse trajectories, and the most rewarding one is selected using a reward prediction model.



\begin{table*}[t!]
    \centering
    \small
    \resizebox{\textwidth}{!}{%
    \begin{tabular}{c|c|c|c|c|c}
        \toprule
        Approach & Name & Model & Parameters & Score & Success Rate \\
        \hline
        \multirow{3}{*}{Zero Shot} & Random  & - & - & 33.74 & 6.80 \\
        & Rule$_{ws}$ $^1$ & - & - & 45.60 & 9.60 \\
        & ZSL-Flan-T5 & Flan-T5-large & 780 Million & 41.10 & 10.30 \\
        \hline
        \multirow{3}{*}{In-context Learning} & Act $^2$  & PaLM & 540 Billion & 62.30 & 30.10 \\
        & ASH $^4$ & CODE-DAVINCI-002 & N/A & 56.70 & 30.20 \\
        & ReAct $^2$ & PaLM & 540 Billion & 66.60 & 40.00 \\
        & AskAct $^3$ & Llama-2 & 70 Billion & 68.60 & 42.20 \\
        \hline
        \multirow{5}{*}{RL-based Method} & PG$_{ws}$ $^1$ & BART, BERT & 516 Million & 52.50 & 11.20 \\
        & DRRN & GRU & 1.2 Million & 46.87 & 11.73 \\
        & RNN$_{ws}$ $^1$ & GRU & 5 Million & 55.20 & 17.60 \\
        & PPO$_{500K}$ (Ours) & Flan-T5-large & 780 Million & 68.19 & 38.55 \\
        & PPO$_{1M}$ (Ours)& Flan-T5-large & 780 Million & \textbf{72.13} & \textbf{42.55} \\
        \hline
        Human & Human $^1$ & - & - & 82.10 & 59.60 \\
        \bottomrule
        \multicolumn{6}{l}{\scriptsize{Results are taken from published research: $^1$ from~\cite{NEURIPS2022_82ad13ec}, $^2$ from~\cite{yao2023react}, 
$^3$ from~\cite{kim2023prospector}, and $^4$ from~\cite{sridhar2023hierarchical}.}
}
    \end{tabular}
    }
\vspace{-4pt}
\caption{Results from methods in the WebShop environment that do not rely on human demonstration data.}
\vspace{2pt}
\label{unsupervised_methods}
\end{table*}

\section{Results}

\subsection{Quantitative Analysis}

\begin{table*}[t!]
    \centering
    \resizebox{\textwidth}{!}{%
    \begin{tabular}{c|c|c|c|c|c}
        \toprule
        Approach & Name & Model & Parameters & Score & Success Rate \\
        \hline
        \multirow{4}{*}{Behavioral Cloning} & PIX2ACT $^3$ & Pix2Struct & 282 Million  & 46.70 & NR \\
        & BC$_{ws}$ $^1$ & BART, BERT & 516 Million  & 59.90 & 29.10 \\
        & BC$_{our}$ & Flan-T5-large & 780 Million & 66.56 & 37.05 \\
        & WebGUM $^2$ & Flan-T5-XL & 3 Billion & 67.50 & 45.00 \\
        \hline
        \multirow{3}{*}{Hybrid Methods} & BC + PG $^1$  & BART, BERT & 516 Million & 62.40 & 28.70 \\
        & AskAct + TR (Prospector) $^4$  & Llama-2, FLAN-T5-XL & 70 + 3 Billion & 70.20 & 43.60 \\
        & BC + PPO$_{500K}$ ({\ours}$_{500K}$) & Flan-T5-large & 780 Million & 74.60 & 46.95 \\
        & BC + PPO$_{1M}$ ({\ours}$_{1M}$) & Flan-T5-large & 780 Million & \textbf{76.87} & \textbf{49.60} \\
        \bottomrule
        \multicolumn{6}{l}{\scriptsize{Results are taken from published research: $^1$ from~\cite{NEURIPS2022_82ad13ec}, $^2$ from~\cite{furuta2024multimodal}, 
$^3$ from~\cite{shaw2023pixels}, and $^4$ from~\cite{kim2023prospector}.}}
    \end{tabular}
    }
\vspace{-4pt}
\caption{Results from methods in the WebShop environment that use human demonstration data.}
\vspace{-6pt}
\label{human_demonstration_methods}
\end{table*}

\stitle{RQ1: Effectiveness of Unsupervised Learning.}
In Table~\ref{unsupervised_methods}, we systematically evaluate the performance of various methods that do not use human demonstrations for training.
Starting with RL-based models, our PPO-trained model with 1 million steps (PPO$_{1M}$) emerges as the top performer, achieving a statistically significant score of 72.13 and a success rate of 42.55. 
Notably, these results surpass those obtained by alternative RL-based approaches, namely PG${ws}$, DRRN, and RNN$_{ws}$, underscoring the superior efficacy of the PPO methodology.
Among In-context learning methods, the AskAct stands out with the most impressive results. 
However, even the best-performing AskAct, 70 billion parameters, fails to outperform a smaller model fine-tuned in an unsupervised setting with PPO (PPO$_{1M}$).
Specifically, in terms of percentage improvements, the PPO-trained model with 1 million steps (PPO$_{1M}$) outperforms the AskAct by 5.15\% on the score metric and approximately 0.83\% on the success rate metric.
This pattern persists when comparing ReAct (540 billion parameters) with PPO$_{1M}$ model. 
This observation suggests that fine-tuning of small models using RL can yield superior performance compared to in-context learning methods.
In addition to RL-based and in-context learning methods, Table~\ref{unsupervised_methods} includes zero-shot learning methods, including zero-shot Flan-T5 (ZSL-Flan-T5) to quantify the role of unsupervised training.
\stitle{RQ2: Impact of Human Demonstrations.}
Table~\ref{human_demonstration_methods} presents the results of various methods incorporating human demonstration.
In the behavioral cloning approach, WebGum emerges as the top performer, leveraging the Flan-T5-XL model with 3 billion parameters. It achieves a score of 67.5 and a success rate of 45.0. 
We also present the results of our fine-tuned Flan-T5-large model (BC$_{our}$) with 780 million parameters. 
Both models outperform the PIX2ACT and BC${ws}$ models, which utilize BART and BERT architectures. 
This notable superiority underscores the effectiveness of instruction-finetuned language models.
Turning to hybrid methods, 
{\ours}$_{500K}$, {\ours}$_{1M}$, and BC + PG models initially undergo refinement through human demonstrations in a supervised setting, followed by additional fine-tuning in an unsupervised setting using RL. 
In contrast, Prospector employs the AskAct method (in-context learning) and a reward prediction model, choosing the most rewarding trajectory through supervised learning. 
Among these approaches, {\ours}$_{1M}$ achieves remarkable performance.
It attains an exceptional Score of 76.87 and a Success Rate of 49.6. 
Notably, our approach surpasses all other hybrid and behavioral cloning methods in both metrics.

\begin{table*}[t!]
    \centering
    \resizebox{\textwidth}{!}{%
    \begin{tabular}{c|c|c|c|c|c}
        \toprule
        Approach & Name & Model & Parameters & Score & Success Rate \\
        \hline
        RL-based Method & PPO$_{1M}$ & Flan-T5-large & 780 Million & 72.12 & 42.55 \\
        Hybrid Method & BC + PPO$_{1M}$ ({\ours}$_{1M}$) & Flan-T5-large & 780 Million & \textbf{76.87} & \underline{49.6} \\
        Unsupervised Domain Adaptation & UDA$_{1M}$ & Flan-T5-large & 780 Million & 74.69 & 46.42 \\
        State-Space Exploration & LASER\cite{ma2023laser} & GPT-4-0613 & N/A & \underline{75.6} & \textbf{50.0} \\
        \bottomrule
    \end{tabular}
    }
    \caption{Comparison of the Best Models.}
    \label{best_methods}
\end{table*}

\begin{table*}[t!]
    \centering
    \resizebox{\textwidth}{!}{%
    \begin{tabular}{c|c|c|c|c|c|c}
        \toprule
         Approach $\xrightarrow[]{}$ & \multicolumn{2}{c|}{Single Domain Behavioral Cloning} & \multicolumn{4}{c}{Unsupervised Domain Adaptation} \\
        \hline
         PPO Adaptation Configs $\xrightarrow[]{}$ & \multicolumn{2}{c|}{No PPO (SDBC)} & \multicolumn{2}{c|}{PPO for 500k steps (UDA$_{500K}$)} & \multicolumn{2}{c}{PPO for 1M steps (UDA$_{1M}$)} \\
        \hline
        Single-domain Supervision $\downarrow$ & Score & Success Rate & Score & Success Rate & Score & Success Rate \\
        \hline
        Fine-tuned on Beauty & 64.23 & 31.41 & 73.99 & 45.80 & 74.49 & 45.85 \\
        Fine-tuned on Garden & 64.79 & 34.76 & 73.97 & 44.70 & 75.27 & 47.5 \\
        Fine-tuned on Grocery & 61.80 & 27.50 & 73.83 & 45.75 & 74.91 & 47.60 \\
        Fine-tuned on Electronics & 62.03 & 30.97 & 73.46 & 45.25 & 74.41 & 44.5 \\
        Fine-tuned on Fashion & 62.54 & 31.60 & 73.37 & 44.45 & 74.36 & 46.65 \\
       \bottomrule
        Average $\xrightarrow[]{}$ & 63.07 & 31.24 & 73.72 & 45.19 & 74.68 & 46.42 \\
        \hline
    \end{tabular}
    }
    \caption{The results of unsupervised domain adaptation and single domain methods in the WebShop environment.}
    \label{UDA_methods}
    \vspace{-10pt}
\end{table*}

\noindent
\underline{Effective Utilization of Human Demonstrations:}
In comparing two variants of the Flan-T5-large model, as presented in Table~\ref{human_demonstration_methods} and Table~\ref{unsupervised_methods}, we focused on one fine-tuned in a supervised setting with human demonstrations (referred to as BC$_{our}$ in Table~\ref{human_demonstration_methods}) and another fine-tuned exclusively with PPO for 1 million steps in an unsupervised setting (referred to as PPO$_{1M}$ in Table~\ref{unsupervised_methods}). 
Surprisingly, the unsupervised model (PPO$_{1M}$) demonstrated an 8.36\% higher score and a 14.84\% higher success rate compared to the supervised model, which is statistically significant.
\emph{This outcome suggests that relying only on human demonstrations does not always lead to superior results.} 
Moreover, when the supervised model is subjected to further training with PPO, it produces the best results.

\noindent 
\underline{Comparison between the Best Models:}
We present the results from the best models in Table~\ref{best_methods}. 
Notably, {\ours}$_{1M}$ achieves a state-of-the-art score (\ie 76.87) surpassing all other models.
Surprisingly, our model, based on Flan-T5-Large (780 million parameters), has outperformed the LASER method, which relies on the latest GPT-4 model with extensive handcrafted prompt, in terms of the Score metric. 
It also achieves comparable performance in terms of Success Rate (49.6 vs 50.0).
These findings strongly suggest that a model, when further fine-tuned with PPO after supervised training, can deliver superior results, even with a relatively smaller model size.

\stitle{RQ3: Unsupervised Domain Adaptation.}
The Single Domain Behavioral Cloning (SDBC) approach involves fine-tuning a Flan-T5-large model in a supervised setting using demonstrations specific to a particular domain (\eg Beauty).
Subsequently, without any additional refinement for other domains, the model is directly tested using the WebShop environment encompassing all domains.
In contrast, UDA takes the Flan-T5-large model fine-tuned in a single domain and further refines it across all domains using PPO in the unsupervised setting.
Table~\ref{UDA_methods} presents two versions of UDA: UDA$_{500K}$ and UDA$_{1M}$.
Both UDA methods exhibit superior performance (\ie statistically significant) in terms of Score and Success Rate metrics when compared to the corresponding metrics of SDBC. 
This superiority is evident not only on a domain-specific basis but also on the average performance across domains.
In particular, concerning the average performance across domains, UDA$_{1M}$ surpasses SDBC by 18.4\% in the Score and 48.6\% in the Success Rate metrics. This emphasizes the crucial role of unsupervised PPO refinement and its impact on enhancing overall performance.

\noindent
\underline{Role of Supervision in a Single Domain:}
To compare the UDA results with RL-based ones, we can refer to Table~\ref{UDA_methods} and Table~\ref{unsupervised_methods}, where UDA$_{500K}$ model outperforms the PPO$_{500K}$ in terms of both Score and Success Rate metrics. 
Similarly, UDA$_{1M}$ surpassed PPO$_{1M}$. 
Specifically, the UDA$_{1M}$ model achieves a 3.5\% higher Score and a 9.09\% higher Success Rate compared to the PPO$_{1M}$ model.
Likewise, the UDA$_{500K}$ model attained an 8.1\% higher Score and a 17.2\% higher Success Rate compared to the PPO$_{500K}$ model.
These findings indicate that incorporating single-domain human demonstration supervision significantly enhances the model's capacity for more effective fine-tuning during unsupervised training with PPO. 
This approach outperforms models that lack any supervised training, which highlights the value of leveraging human demonstrations in the adaptation process.

\begin{figure}[t]
  \centering
  \vspace{-20pt}
  \includegraphics[width=0.5\textwidth]
  {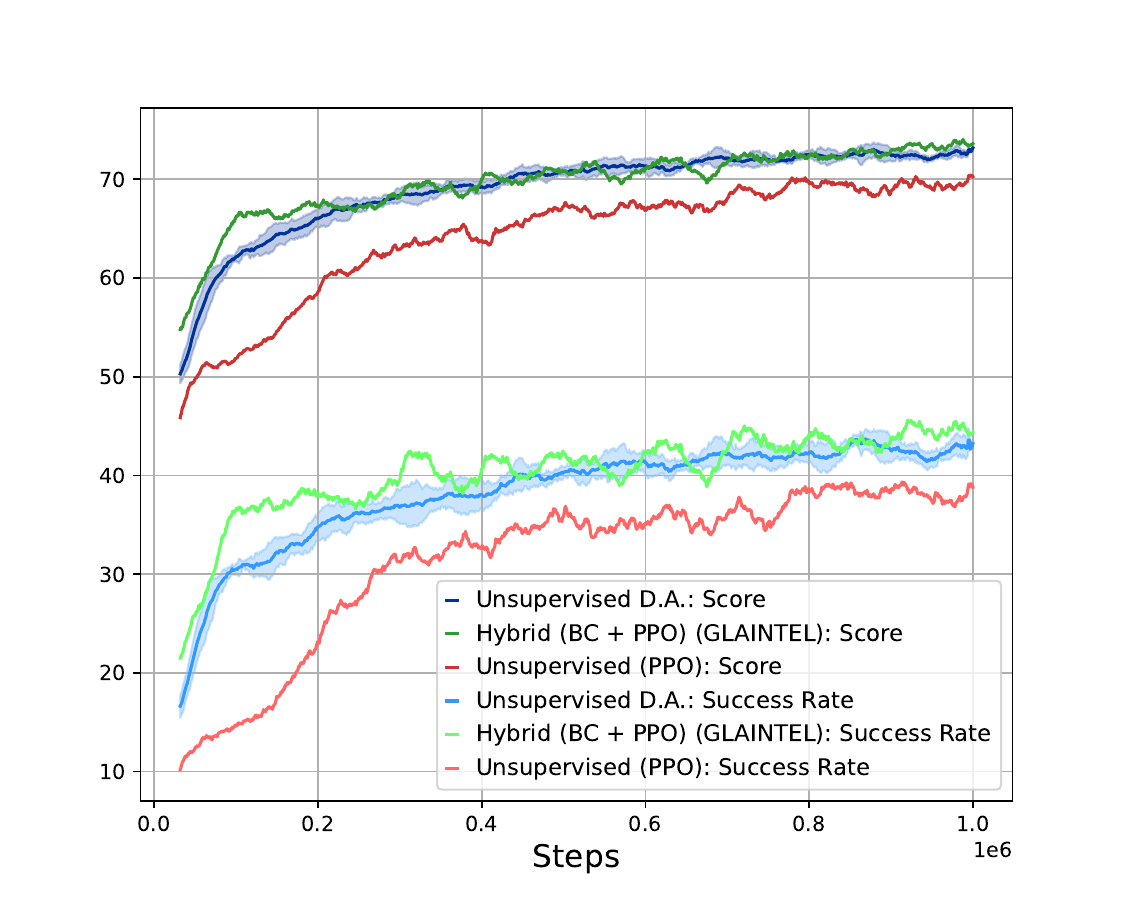}
  \vspace{-14pt}
  \caption{Learning curves of different methodologies: Unsupervised Domain Adaptation (UDA), Hybrid (BC + PPO) ({\ours}), and RL-based Unsupervised (PPO).}
  \label{fig:DifferentTraining}
  \vspace{-14pt}
\end{figure}

\stitle{Learning Curves for PPO training.} In Figure~\ref{fig:DifferentTraining}, the learning curves of Score and Success Rate metrics during PPO fine-tuning are illustrated for various methodologies: the UDA, the hybrid ({\ours}) (BC + PPO), and the RL-based PPO. 
Both the hybrid method and the unsupervised domain adaptation method demonstrate higher sample efficiency compared to the unsupervised method. 
This aligns with expectations, considering that both the hybrid method and the unsupervised domain adaptation method underwent some level of supervised training before RL fine-tuning -- a contrast to the RL-based unsupervised method, which did not.

\begin{table*}[t!]
    \centering
    \small
    \resizebox{\textwidth}{!}{%
    \begin{tabular}{c|c|c|c|c|c}
        \toprule
        Approach & Name & Model & Parameters & Score & Success Rate \\
        \hline
        Hybrid Method & BC + PG & BART, BERT & 516 Million & 59.25 & 24 \\
        Hybrid Method & BC + PPO$_{1M}$ ({\ours}$_{1M}$) & Flan-T5-large & 780 Million & 78.35 & 53 \\
        State-Space Exploration & LASER & GPT-4-0613 & N/A & 83.55 & 56 \\
        \bottomrule
    \end{tabular}
    }
    \caption{Results of Zero-shot simulation-to-real experiment on eBay.}
    \label{Sim_to_real}
\end{table*}

\subsection{Results on Real Website: eBay}
We also conduct limited evaluations on a real website: eBay. 
For this experiment, we evaluate the performance of three methods: \myNum{i}~our best model ({\ours}$_{1M}$), \myNum{ii}~the GPT-4-based method LASER, and \myNum{iii}~the WebShop baseline (BC + PG).
\emph{It is important to highlight that we used the models trained using the Webshop environment and did not perform any fine-tuning using the eBay website.}
Following~\cite{NEURIPS2022_82ad13ec}, we randomly sampled 100 user instructions to evaluate the performance of these methods. 
As presented in Table \ref{Sim_to_real}, our method {\ours}$_{1M}$ significantly outperformed the WebShop baseline (BC + PG) by 32.23\% in the Score metric and by 120.83\% in the Success Rate metric. 
Moreover, although LASER, utilizing GPT-4, has slightly higher Score and Success Rate metrics compared to our model {\ours}$_{1M}$, we are confident that {\ours}$_{1M}$ can achieve comparable or even superior results by enabling of unsupervised training using PPO. 
Additionally, it is worth noting that our approach utilizes a 780 million parameter model, which is significantly smaller than GPT-4, not to mention the costs associated with GPT-4.
\emph{
We present an ablation study in Appendix~\ref{sec:ablation_study}.}

\section{Related Work}

\stitle{Fine-tuning LLMs with RL and Human Feedback.}
Fine-tuning LLMs with human feedback and reinforcement learning has been studied extensively. 
\cite{nakano2021webgpt} developed the WebGPT by fine-tuning the GPT-3 model using behavior cloning and rejection sampling. 
Moreover, InstructGPT~\cite{ouyang2022training} was developed using the three-step approach: supervised fine-tuning, reward model training, and reinforcement learning via PPO with the help of the trained reward model. 
Additionally, the authors in~\cite{stiennon2020learning} fine-tuned a model that may choose a human-preferred summary, they used this model as a reward function to fine-tune a summarization policy using RL.

\stitle{Foundation Models for Decision Making.} Foundation models possess robust decision-making capabilities, rendering them invaluable across various downstream tasks. 
For instance, recent works~\cite{ahn2022can,pmlr-v162-huang22a, huang2022inner} showcase the application of foundation models in the robotics domain. 
Moreover, works~\cite{rawles2023android,Wen2023EmpoweringLT,Yan2023GPT4VIW,hong2023cogagent} utilize foundation models to intelligently navigate Android applications.
Additionally, the foundation models have been utilized in gaming contexts~\cite{doi:10.1126/science.ade9097, NEURIPS2022_b2cac94f, reed2022generalist, fan2022minedojo, wang2023voyager, carta2023grounding}.

\stitle{Web Navigation.}
Many benchmarks and datasets exist for the training and assessment of web agents \cite{NEURIPS2022_82ad13ec, shi2017world, deng2023mind2web, zhou2023webarena, liu2018reinforcement}. 
Researchers have consequently proposed diverse web agents and tested their performance on these benchmarks. 
The MiniWob++ benchmark is among these benchmarks on which different methods have been applied. 
For example, \cite{Humphreys2022ADA} employed a combination of reinforcement learning and behavioral cloning, \cite{furuta2024multimodal} utilized supervised training on an instruction-fine-tuned LLM, \cite{liu2018reinforcement} introduced Workflow-guided exploration (WGE), and \cite{gur2018learning} trained DQN agents (QWeb network and INET network). 
Additionally, the Mind2Web benchmark introduced the MindAct model, synergizing the strength of small and large LLMs~\cite{deng2023mind2web}. 
Additionally, a visual language model named CogAgent was utilized for the benchmark~\cite{hong2023cogagent}. 
\cite{zeng2023agenttuning} presented AgentTuning as another notable approach to tackle the Mind2Web benchmark. 
Furthermore, considering the Webshop benchmark, various methodologies have been proposed that use in-context learning ~\cite{kim2023prospector, yao2023react, sridhar2023hierarchical}, supervised learning~\cite{furuta2024multimodal, shaw2023pixels}, and RL~\cite{NEURIPS2022_82ad13ec}.
\emph{Nonetheless, no work has clearly outlined the impact of human demonstrations and the optimal utilization of available demonstration data. 
Furthermore, UDA remains underexplored.}

\section{Conclusion}
We introduce {\ours}, a flexible agent designed for training across diverse product search scenarios, accommodating situations with limited or no human demonstrations for supervision. 
We also investigate the optimal utilization of demonstration data, showing that straightforward supervised learning approaches, like behavior cloning, do not yield superior results when using human demonstration data. 
Through extensive experimental evaluations in the WebShop environment, we highlight the crucial role of the unsupervised training phase employing the PPO algorithm. 
When combined with supervised learning, this approach achieved results comparable to methods utilizing GPT-4. 
Additionally, we explore an underexplored scenario where demonstration data is confined to a single domain, we employ UDA techniques to accommodate novel domains.
We also present evaluations on a real website, eBay, to showcase the applicability of {\ours} in the real world.


\section*{Acknowledgments}
This work is supported in part by the National Science Foundation~(NSF) under grant IIS-2401685.

\section{Limitations}
In our experiments, we only used the current and previous observations as input to the model. Although including additional observations (\eg the last four observations) can potentially improve performance, it is important to consider that the increase in the number of observations also expands the size of the context, leading to requirements for higher GPU memory.
Moreover, the current architecture relies only on textual descriptions of the environment, without embedding screenshots of web pages or product images. Improving the performance of the agent can be achieved by integrating these visual elements into the model.

It should be noted that other web environments, such as MiniWoB~\cite{shi2017world}, have simple, plain backgrounds and minimal interaction within a small area of 160 x 160 pixels. Because of these limitations, we did not assess our method in this environment and considered a more realistic environment, WebShop. However, we plan to evaluate the performance of our approach in other web environments in the future.

\bibliography{custom}


\appendix

\section{Implementation Details}
\label{sec:implementation_details}

\begin{table}[t]
    \centering
    \begin{tabular}{c|c}
        \toprule
        Hyperparameter & Value \\
        \hline
        Number of Epochs & 10 \\
        Learning Rate & $2 \times 10^{-5}$ \\
        Warmup Steps & 100 \\
        Weight Decay & 0.01  \\
        Batch Size & 32 \\
        Adam Optimizer Epsilon & $10^{-8}$ \\ 
        Adam Optimizer $\beta_1$ & 0.9 \\
        Adam Optimizer $\beta_2$ & 0.999 \\ 
        \bottomrule
    \end{tabular}
    \caption{Supervised Learning Hyperparameters.}
    \label{SL_hyperparameters}
\end{table}

\begin{table}[t]
    \centering
    \resizebox{\columnwidth}{!}{%
    \begin{tabular}{c|c}
        \toprule
        Hyperparameter & Value \\
        \hline
        \# of collected transitions \\between two updates & 640 (16 × 40) \\
        Number of epochs per update & 1 \\ 
        Batch Size & 8 \\
        Learning Rate & $10^{-6}$ \\
        Adam Optimizer Epsilon & $10^{-5}$ \\ 
        Adam Optimizer $\beta_1$ & 0.9 \\
        Adam Optimizer $\beta_2$ & 0.999 \\ 
        Discount Factor & 0.99 \\
        Lambda for Generalized \\Advantage Estimate & 0.99 \\
        Entropy Loss Coefficient & 0.01 \\
        Value Loss Coefficient & 0.5 \\
        Maximum Gradient Norm & 0.5 \\
        Clipping Epsilon & 0.2 \\
        \bottomrule
    \end{tabular}
    }
    \caption{Unsupervised Learning Hyperparameters.}
    \label{PPO_hyperparameters}
\end{table}

Our implementation operates on a client-server architecture, with the training scripts serving as the client and communicating requests to LLM servers.
Specifically, a master server manages these requests, distributing them across multiple LLM servers.
Once each LLM server completes its computations, the master server consolidates the results and sends them back to the training script.
Furthermore, we use vertical model parallelism, enabling the parallelization of individual LLMs across multiple GPUs.
In our experiments, we utilized a single LLM, Flan-T5-Large, with 780 million parameters.
This model was parallelized across 4 Nvidia V100 32GB GPUs.
We incorporated the last two observations as the model input and an encoder context size of 1024.

To train the agent using the human demonstrations, we used the Trainer library provided by Huggingface \footnote{Trainer: \href{https://huggingface.co/docs/transformers/main_classes/trainer}{https://huggingface.co/docs/transformers /main\_classes/trainer}}. 
We employed the Adam optimizer, and for the remaining hyperparameter values, refer to Table~\ref{SL_hyperparameters}.
In our unsupervised learning phase, we leverage the PPO algorithm, and the complete values of hyperparameters can be found in Table~\ref{PPO_hyperparameters}.

\section{Ablation Study}
\label{sec:ablation_study}

\begin{figure}[t!]
    \centering
    \includegraphics[width=\linewidth]{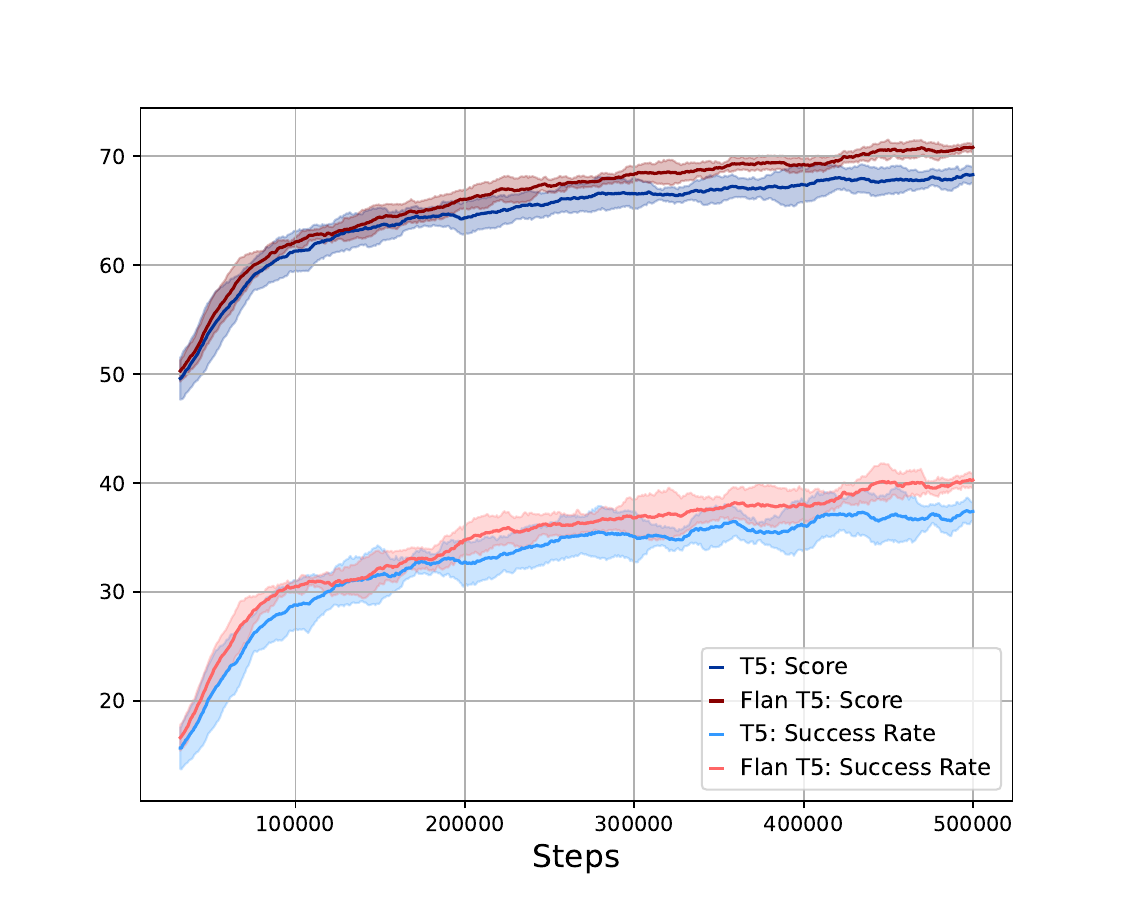}
 \caption{Hybrid setting: BC + PPO: Flan-T5 is more sample efficient than T5 model.}
 \label{fig:T5VsFlanT5}
\end{figure}

\stitle{Flan-T5 vs T5.}
We employed two models of identical size, each with 780 million parameters: Flan-T5-Large and T5-Large.
The results, as presented in Table~\ref{t5_flan_t5}, demonstrate that adopting the Flan-T5-Large model instead of T5-Large leads to a substantial improvement of 30.93\% in the Score and a remarkable 52.07\% increase in the Success Rate in the unsupervised setting (PPO). 
Furthermore, in the domain adaptation scenario, we observed a 2.60\% Score enhancement and a 4.85\% improvement in the Success Rate. Moreover, 
Figure~\ref{fig:T5VsFlanT5} demonstrates that employing the Flan-T5 model over the T5 model results in better sample efficiency. 
Specifically, both Score and Success Rate metrics exhibit faster growth during PPO fine-tuning in the Flan-T5 model compared to the T5 model. 
This outcome was anticipated as the Flan-T5 model enjoys the advantage of being fine-tuned on user instructions, a benefit not shared by the T5 model.

\begin{table}[t]
    \centering
    
    \resizebox{\columnwidth}{!}{%
    \begin{tabular}{c|c|c|c|c}
        \toprule
        Configs $\xrightarrow[]{}$  & \multicolumn{2}{c|}{SL (one cat) + PPO (500k)} & \multicolumn{2}{c}{PPO (500k)}\\
        \hline
        Model $\downarrow$ &Score&Success Rate&Score&Success Rate\\
        \hline
        Flan-T5 & \textbf{73.72} & \textbf{45.19}  & \textbf{68.18} & \textbf{38.55}\\
        \hline
        T5 & 71.85 & 43.10 & 52.07 & 25.35 \\
        \bottomrule
    \end{tabular}
    }
    \caption{Ablation Study (T5 vs Flan-T5)}
    \label{t5_flan_t5}
\end{table}

\begin{figure}[t!]
  \centering
  \vspace{-20pt}
  \includegraphics[width=0.46\textwidth]
  {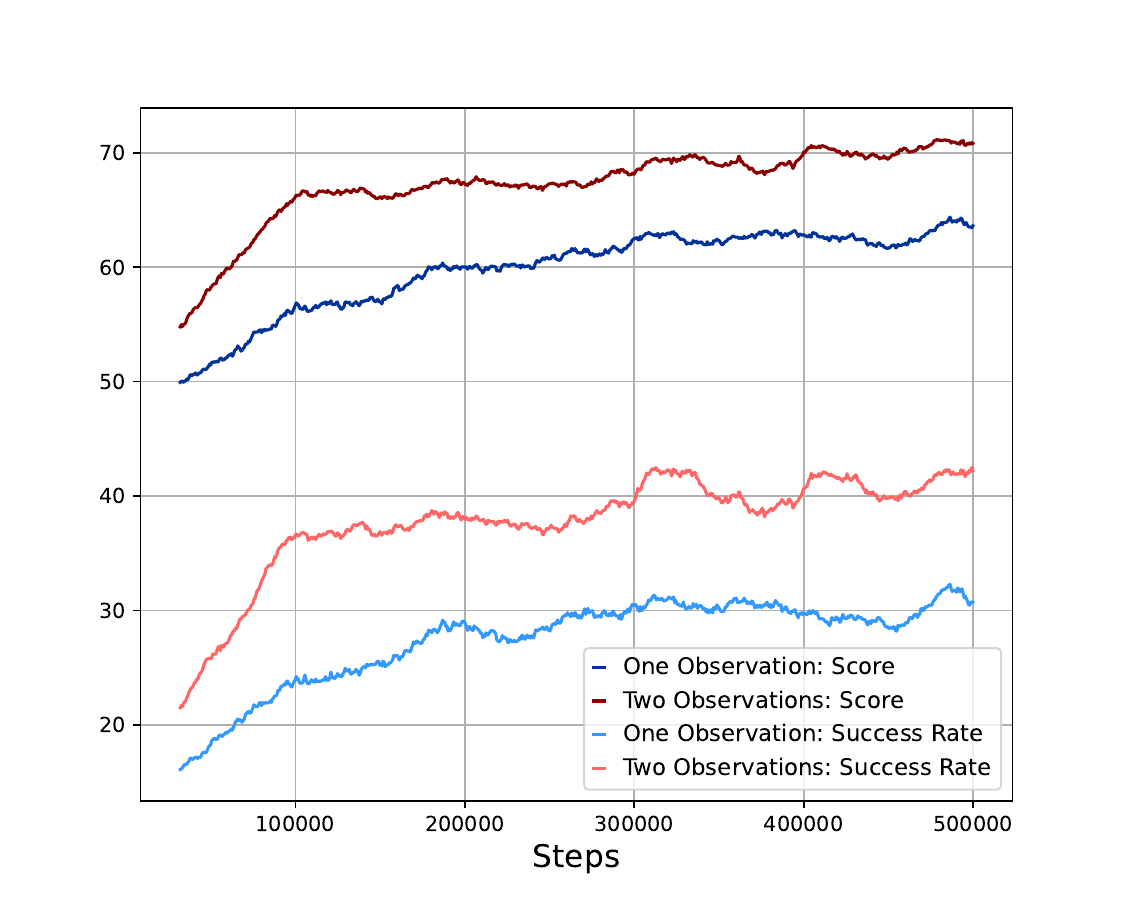}
  \vspace{-2pt}
  \caption{The model is more sample efficient when we feed it with the last two observations.}
  \label{fig:2obsVs1obs}
  \vspace{-2pt}
\end{figure}

\stitle{2 Observations vs 1 Observation.}
As demonstrated in Table \ref{2_obs_vs_1_obs}, combining the present observation state with the preceding observation state to create a historical context and subsequently providing the model with this new observation containing both leads to a notable 10.54\% boost in the Score and a remarkable 36.21\% improvement in Success Rate in the supervised setting. 
This substantial enhancement is equally observable in the context of the hybrid method (SL + PPO) where the supervised training is coupled with unsupervised training (PPO), resulting in a significant 14.26\% increase in the Score and an impressive 39.73\% improvement in Success Rate. 
Additionally, during the training, we noticed that employing a historical context (having the current and last observations) as input enhances the sample efficiency for the agent compared to using just one observation (see Figure~\ref{fig:2obsVs1obs}). 
Specifically, Score and Success Rate metrics show a swifter increase with fewer steps when leveraging two observations (historical context) as input, while the progression is notably slower when utilizing only a single (or current) observation.

\begin{table}[t!]
    \centering
    
    \resizebox{\columnwidth}{!}{%
    \begin{tabular}{c|c|c|c|c}
        \toprule
        Configs $\xrightarrow[]{}$ & \multicolumn{2}{c|}{SL (all cats)} & \multicolumn{2}{c}{SL + PPO (500k)}\\
        \hline
          & Score&Success Rate&Score&Success Rate\\
        \hline
        2 observations & \textbf{66.55} & \textbf{37.05} & \textbf{74.60} & \textbf{46.95}\\
        \hline
        1 observation & 60.20 & 27.20 & 65.29 & 33.60 \\
        \bottomrule
    \end{tabular}
    }
    \caption{Ablation Study (2 observations vs 1 observation)}
    \label{2_obs_vs_1_obs}
    \vspace{-2pt}
\end{table}

\vspace{4pt}
\stitle{Comparison of Decoding Methods.}
In Table \ref{inference_methods}, we compare the performance of four different decoding methods: \myNum{i}~Epsioln-Greedy algorithm (with epsilon value of 0.2), \myNum{ii}~Sampling with top\_p (with top\_p = 0.8 and top\_k = 0.0),\myNum{iii}~Sampling with no top\_p and no top\_k, and \myNum{iv}~Argmax. 
These results are determined by averaging the results achieved from models trained with different techniques and settings, including RL and UDA, among others. 
These results show that, on average, the Epsilon-Greedy algorithm consistently attains the best results during inference, with a Score of 68.23 and a Success Rate of 39.29. 
Following closely, the nucleus sampling (top\_p) method has lower Scores and Success Rates of 66.25 and 37.32, respectively. 
In the third position, traditional sampling produces a score of 65.92 and a Success Rate of 36.41. 
The worst outcomes are associated with the Argmax method, primarily since Argmax frequently causes the web agent to become stuck in a loop. 
In simpler terms, the web agent ends up repeatedly navigating back and forth between web pages.

\begin{table}[t!]
    \centering
    \resizebox{\columnwidth}{!}{%
    \begin{tabular}{c|c|c}
        \toprule
        Comparison & Score & Success Rate\\
        \hline
        Epsilon-Greedy algorithm & \textbf{68.23} & \textbf{39.29} \\
        \hline
        Sampling with top\_p & 66.25 & 37.32 \\
        \hline
        Sampling & 65.92 & 36.41 \\
        \hline
        Argmax & 57.92 & 35.59 \\
        \bottomrule
    \end{tabular}
    }
    \vspace{-2pt}
    \caption{Ablation Study (Decoding Methods)} 
    \label{inference_methods}
    \vspace{-2pt}
\end{table}





\end{document}